# Error Estimation in Approximate Bayesian Belief Network Inference

**Enrique F. Castillo**
Applied Mathematics Dept
University of Cantabria
39005 Santander, Spain
castie@ccaix3.unican.es

**Remco R. Bouckaert**
Computer Science Dept.
Utrecht University
The Netherlands
remco@cs.ruu.nl

**José M. Sarabia**
Economics Dept.
University of Cantabria
39005 Santander, Spain

**Cristina Solares**
Applied Mathematics Dept.
University of Cantabria
39005 Santander, Spain

## Abstract

We can perform inference in Bayesian be-
lief networks by enumerating instantiations
with high probability thus approximating
the marginals. In this paper, we present a
method for determining the fraction of in-
stantiations that has to be considered such
that the absolute error in the marginals does
not exceed a predefined value. The method
is based on extreme value theory. Essentially,
the proposed method uses the reversed gener-
alized Pareto distribution to model probabili-
ties of instantiations below a given threshold.
Based on this distribution, an estimate of the
maximal absolute error if instantiations with
probability smaller than $u$ are disregarded
can be made.

## 1 INTRODUCTION

When propagating uncertainty in Bayesian networks,
one has to account for the probabilities of certain
events. In general, most events have a very small prob-
ability of occurrence and the sum of the probabilities
of rare events is negligible. Many techniques applied
to Bayesian belief networks rely on this assumption.
For example, various approximation algorithms for in-
ference [5, 13, 16, 15] and caching of probabilities of
likely events [14] are founded on this assumption. In
this paper we concentrate on the error obtained in ap-
proximating marginal probabilities.

However, on fore-hand it is not known what the to-
tal contribution of the events with low probability to
the marginals is. As a result, it is difficult to speak
in terms of absolute error-bounds. On the other hand,
if we know on fore-hand what the total contribution
of the events with probability smaller than $q$ is to the
marginals, we can calculate, for example, the number
of iterations of simulation algorithms when a certain
error-bound is demanded. A variant of stratified simu-
lation [3, 4] is guaranteed to visit all events with prob-
ability larger than $1/m$ where $m$ is the number of it-

erations. Furthermore, we can determine a reasonable
size of a cache as proposed in [14].

Druzdzel [12] gave a solution for the problem by defin-
ing a random variable $X$ as the logarithm of the proba-
bilities of an event and selecting events with a uniform
distribution. He showed that under some general con-
ditions the distribution over $X$ can be approximated
by a log-normal distribution. From this distribution,
an estimate can be made of the total contribution
of the events with probability smaller than $q$ to the
marginals.

However, the log-normal approximation is based on
the central limit theorem. As a result, the approxima-
tion is good in the neighborhood of the mean of $X$.
But we are interested in the tails of the distribution
over $X$, where the log-normal approximation can dif-
ferentiate considerably from the real distribution. In
fact, estimates based on the log-normal approximation
differ so much that reliable estimates are not possible
for the tails.

An alternative for the log-normal approximation is ex-
treme value theory [6]. This theory is engaged with
the tails (right or left) of distributions. In this paper,
we define a random variable $X$ as the probabilities
of events and select events with a distribution propor-
tional to the probability of their occurrence. Note that
we do not take the logarithm as [12] does, though we
can work with logarithms as well with a small change
in the interpretation of our model. We show how to
approximate the left tail of the distribution over $X$,
which gives us the total contribution of the events with
probability smaller than $q$ to the marginals. Our the-
ory applies to distributions involving continuous vari-
ables and gives a very good approximation for discrete
variables.

In Section 2 we give a formal statement of the prob-
lem and some definitions. In Section 3 we present our
model for solving the problem and in Section 4 we show
how to estimate the various parameters of the model.
We performed some experiments to get insight in the
usability of our method. The results are presented in
Section 5. Finally, in Section 6 we make some conclud-
ing remarks.

## 2    STATEMENT OF THE PROBLEM

A *Bayesian belief network* $B$ over a set of variables $V = \{x_1, \ldots, x_n\}$ is a pair $(B_S, B_P)$, where $B_S$ is a directed acyclic graph over $V$ and $B_P$ is the set of conditional probabilities of $X_i$ given its parents $\pi_i$. A Bayesian belief network defines a probability distribution [17] over $V$ $P_B(V) = \prod_{i=1}^{n} P(x_i|\pi_i)$.

Inference in such knowledge based systems over $V$ consists of the calculations for each $x_i \in V$ of the marginals $P(x_i|E) = \sum_{x_j \in V \backslash Ex_i} P_B(V|E)$ where values of certain variables $E \subset V$ are known to have values $e_i$ for $x_i \in E$. For ease of exposition, we assume that $E = \emptyset$, that is, that there is no evidence. Since the above summation is computational infeasible when $n$ is large, the marginal can be approximated by summing over a subset of instantiations of $V$ with high probability. We are interested in calculating the error that occurs in such an approximation. More specific, we are interested in determining the contribution of all instantiations with probability smaller than $q$ to the total probability mass,

$$G(q) = \sum_{P(V) < q} P(V).$$

For continuous variables, consider the function $P : I_V \rightarrow I\!R^+$, defined as

$$P(x_1, \ldots, x_n) = \prod_{i=1}^{n} P(x_i|\pi_i), \qquad (1)$$

where $I_V$ is the set of all possible instantiations of the set $V$, and $I\!R^+$ is the set of all non-negative real numbers, that is, we associate with each instantiation its probability density function (pdf) value. Assuming some regularity conditions, $P$ is a unidimensional random variable.

Let $f(p)$ be the probability density function of $P$. The elemental contribution of all instantiations with probability $p \leq P \leq p + dp$ to the total probability mass is $pf(p)dp$. Further, the fraction of the contribution of all instantiations with probability smaller than $q$ to the total probability mass becomes

$$G(q) = \frac{\int_0^q pf(p)dp}{\int_0^\infty pf(p)dp} = \frac{\int_0^{F(q)} F^{-1}(u)du}{\int_0^1 F^{-1}(u)du}, \qquad (2)$$

where the denominator is the expected value of $P$ and we have made the change of variable $u = F(p)$, where $F(p)$ is the cumulative density function (cdf) of $P$, to get the right-hand side expression. Note that $G(q)$ is the Lorenz curve of $P$ ([1, 2, 11, 10]), which is a cdf that can be associated with a new random variable $Q$ with the same domain as $P$.

Alternatively, we can write the Lorenz curve in terms of the proportion $r$ of instantiations contributing a given fraction of the total probability mass, by making the extra change of variable $r = F(q)$ to get

$$L(r) = G(F^{-1}(r)) = \frac{\int_0^r F^{-1}(u)du}{\int_0^1 F^{-1}(u)du}, \qquad (3)$$

where $L(r)$ is a cdf with associated domain $[0, 1]$. However, in this paper we shall use the Lorenz curve (2).

Assume that we consider the subset $V_{p_0}$ of all instantiations with associated $p$ values such that $p \geq p_0$ and we calculate marginal probabilities based on this subset $V_{p_0}$, instead of the set $V$. Then, $G(p_0)$ is an upper bound for the error of the marginal probabilities. Thus, we can take $G$ as the basis for error estimation.

For example, Figure 1 shows both densities $f(p)$ and $g(p)$, where $f(p)$ is the density of $P$ and $g(p)$ the density of the contribution to the total probability mass of instantiations with associated probability $p$. Now 50 % of the instantiations, i.e., instantiations with associated probability $P > 10$, contribute 98.8 % to the total probability mass (the area below the $g$ function in the region $P > 10$). Consequently, consideration of only these instantiations leads to a maximum error of 0.012 in any probability evaluation.

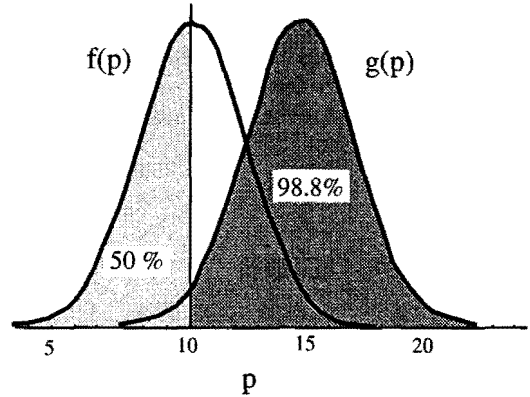

Figure 1: Illustration of how 50 % of the instantiations can contribute 98.8 % to the total probability mass.

The problem that we consider in this paper is the approximation of $G$. More precisely, we try to estimate the left tail of $G$.

## 3    PROPOSED MODEL

Thus, we are not interested in estimating $f(p)$ but the curve $G(q)$ of $P$, or to be more precise its left tail. Due to the fact that we want small errors in the evaluation of marginal probabilities, we consider neglecting a small fraction of the probability mass, that is, we really want to estimate lower percentiles of $G(q)$. Thus, we must deal with the left tail of the associated distribution $G(q)$. This could be achieved by randomly generating a sample for $G$ and straightforward estimation. It is well known that in order to get an estimate of a small proportion $p$ with a given relative error, the required sample size is proportional to $1/p$. Since $p$ is very small in our case, a very large sample would be necessary. An alternative is to use extreme value theory.

To apply extreme value theory, we consider the random variable $Q$ with associated cdf $G(q)$. We transform $Q$ by truncating it at a threshold value $u$ and then translating the origin to $u$, thus getting a new random variable $S$, which is defined as $Q - u$ given $Q \leq u$. The random variable $S$ has the cdf $H(s; u)$ which is

$$Pr(S \leq s) = Pr\left[Q \leq u + s | Q \leq u\right] = \frac{G(u+s)}{G(u)}; \quad (4)$$

for $q_0 - u \leq s \leq 0$, where $G(q)$ is the cdf of $Q$ and $q_0$ is the lower endpoint of $G$ or $F$.

It follows from (4) that $G(q)$ for small values of $p$ can be written as

$$G(q) = G(u)H(q - u; u), q_0 < q < u. \quad (5)$$

Thus, estimating $G(q)$ is equivalent to estimating: (a) $G(u)$ and (b) $H(q-u; u)$. We discuss these two points in the following paragraphs.

To choose $H(q-u; u)$, we use the Reversed Generalized Pareto Distribution (RGPD), which will be justified by Theorem 1. The RGPD $U(z; \delta, \alpha)$ is a distribution defined by

$$U(z; \delta, \alpha) = \begin{cases} \left(1 + \frac{z}{\delta}\right)^{\frac{1}{\alpha}}, & \text{if } \alpha < 0, \quad z < 0, \\ & \text{or } \alpha > 0, \quad -\delta < z < 0, \\ \exp\left(\frac{z}{\delta}\right), & \text{if } \alpha = 0, \quad z < 0, \end{cases} \quad (6)$$

where $\delta$ and $\alpha$ are the scale and shape parameters, respectively.

**Theorem 1** *The RGPD $U(z; \delta, \alpha)$ is a good approximation of $H(q; u)$, in the sense that*

$$\lim_{u \to q_0} \sup_{q_0 - u < q < 0} |H(q; u) - U(q; \delta(u), \alpha)| = 0, \quad (7)$$

*for some fixed $\alpha$ and functions $\delta(u)$, if and only if $H$ is in the minimal domain of attraction of one extreme value distribution.*

Note that practically all cdfs used in textbooks satisfy this condition. See Pickands [18] for a proof of the equivalent result for the maximal domain of attraction. Using Theorem 1, from (5) and (6), the proposed model for the left tail of $Q$ is

$$G(q) = G(u)U(q - u; \delta(u), \alpha); \quad q_0 < q < u, \quad (8)$$

which depends on $G(u)$ and two parameters $\delta$ and $\alpha$ for each threshold value $u$.

## 4  ESTIMATION OF THE MODEL

The main problem for estimating $\delta$ and $\alpha$ is that the associated random variables $S$ and $Q$ are not directly observable. However, we can observe $P$ and obtain an ordered sample $(p_1, \ldots, p_n)$. A natural estimator for $G(q)$ based on this sample is

$$\hat{G}(q) = \frac{\sum_{p_i < q} 1}{\sum_{p_i < u} 1} \quad (9)$$

for the discrete case and

$$\hat{G}(q) = \frac{\sum_{p_i < q} p_i}{\sum_{i=1}^{n} p_i} \quad (10)$$

for the continuous case. To estimate $G(q)$ for $q < u$, one needs both $G(u)$ and $H(q-u; u)$. In practice, different values of $u$ should be tried out and $G(u)$ can be estimated by (9) and (10) for the discrete and continuous case, respectively.

For estimating $\delta$ and $\alpha$ we use the method proposed by Castillo et al. [7, 8, 9].

Let $I = \{i, j\}, i < j \in \{1, \ldots, n\}$, then we have for the discrete case

$$U(p_i - u; \delta(u), \alpha) = \frac{\sum_{p_k < p_i} 1}{\sum_{p_k < u} 1},$$

$$U(p_j - u; \delta(u), \alpha) = \frac{\sum_{p_k < p_j} 1}{\sum_{p_k < u} 1}, \quad (11)$$

and for the continuous case

$$U(p_i - u; \delta(u), \alpha) = \frac{\sum_{p_k < p_i} p_k}{\sum_{p_k < u} p_k},$$

$$U(p_j - u; \delta(u), \alpha) = \frac{\sum_{p_k < p_j} p_k}{\sum_{p_k < u} p_k}. \quad (12)$$

Substituting (6) in both (11) and (12) and taking the logarithm[1], we obtain

$$\log(1 + (p_i - u)/\delta) = \alpha C_i$$
$$\log(1 + (p_j - u)/\delta) = \alpha C_j, \quad (13)$$

where $C_i = \log\left(\frac{\sum_{p_k < p_i} 1}{\sum_{p_k < u} 1}\right)$ and $C_i = \log\left(\frac{\sum_{p_k < p_i} p_k}{\sum_{p_k < u} p_k}\right)$

for the discrete and continuous case, respectively. It can be seen that (13) is a system of two equations in two unknowns, $\delta$ and $\alpha$. Eliminating $\alpha$, we obtain

$$C_i \log(1 + (p_j - u)/\delta) = C_j \log(1 + (p_i - u)/\delta). \quad (14)$$

We now show that the above estimators are well defined, that is, (14) has one more solution in addition to the trivial solutions $\delta = \pm\infty$, $\alpha = 0$.

---

[1] We use the natural logarithm throughout this article.

**Theorem 2** *Equation (14) has a finite solution, in the interval $(\delta_0, 0)$ if $C_i(u - p_j) - C_j(u - p_i) > 0$, or in the interval $(u - p_i, \delta_0)$ if $C_i(u - p_j) - C_j(u - p_i) <= 0$, where*

$$\delta_0 = \frac{(C_i - C_j)(u - p_i)(u - p_j)}{C_i(u - p_j) - C_j(u - p_i)}. \tag{15}$$

A proof of Theorem 2, which is given in the Appendix, is constructive for it gives rise to the following algorithm for solving (14) for $\delta$. Equation (14) is a function of only one variable, hence it can be solved easily using the bisection method as outlined in Theorem 1 and Algorithms I and II below. Thus, using Algorithm I, one can solve (14), for $\delta$ and obtain an estimate of $\delta$, $\hat{\delta}(i, j)$, say. This estimate is then substituted in one of the two equations in (13) to obtain a corresponding estimate of $\alpha$, $\hat{\alpha}(i, j)$, which is given by

$$\hat{\alpha}(i, j) = \log(1 + (p_i - u)/\hat{\delta}(i, j))/C_i. \tag{16}$$

**Algorithm I**$(p, i, j)$ {$p$ is an array with ordered samples and $i$ and $j$ two indices}

1. Compute $C_i$ and $C_j$ and let $d = C_i(u - p_j) - C_j(u - p_i)$

2. **case**

   - $d = 0$: $\hat{\delta}(i, j) = \pm\infty$. This means $\hat{\alpha}(i, j) = 0$.
   - $d < 0$: Use the bisection method on the interval $[u - p_i, \delta_0]$, where $\delta_0 = (C_i - C_j)(u - p_i)(u - p_j)/d$, to obtain a solution $\hat{\delta}(i, j)$ of (14)
   - $d > 0$: Use the bisection method on the interval $[\delta_0, 0]$, where $\delta_0 = (C_i - C_j)(u - p_i)(u - p_j)/d$, to obtain a solution $\hat{\delta}(i, j)$ of (14).

   **end case**

3. Use $\hat{\delta}(i, j)$ to compute $\hat{\alpha}(i, j)$ using (16).

### 4.1    FINAL ESTIMATES

The estimates found by Algorithm I are based on only two order statistics $\{p_{i:n}, p_{j:n}\}$, thus they do not utilize the information contained in other order statistics. Statistically more efficient estimates are obtained using the following algorithm.

**Algorithm II**

1. Use Algorithm I to compute $\hat{\delta}(i, j)$ *and* $\hat{\alpha}(i, j)$, where $i = m/10; j = i+1, \ldots, m-1$ and $m$ is the number of data points below the threshold $u$.

2. Apply a robust function $R(.)$ to each of the above sets of estimates and obtain a corresponding overall estimates of $\delta$ and $\alpha$.

Examples of the robust function, $R(.)$ in Step 2, include the median and the least median of squares (LMS), Rousseeuw [19]. Thus, overall estimates of $\delta$ and $\alpha$ can be defined as

$$\hat{\delta}_{MED} = median(\hat{\delta}(i, i+1), \ldots, \hat{\delta}(i, m-1)),$$
$$\hat{\alpha}_{MED} = median(\hat{\alpha}(i, i+1), \ldots, \hat{\alpha}(i, m-1)), \tag{17}$$

or

$$\hat{\delta}_{LMS} = LMS(\hat{\delta}(i, i+1), \ldots, \hat{\delta}(i, m-1)),$$
$$\hat{\alpha}_{LMS} = LMS(\hat{\alpha}(i, i+1), \ldots, \hat{\alpha}(i, m-1)), \tag{18}$$

where $median(y_1, y_2, \ldots, y_n)$ is the median of $\{y_1, y_2, \ldots, y_n\}$, and $LMS(y_1, y_2, \ldots, y_n)$ is the estimate obtained using the LMS methods, which in this case is equal to the midpoint of the shortest interval containing half of the numbers $y_1, y_2, \ldots, y_n$ (see Rousseeuw and Leroy [20], pp. 169).

## 5    EXPERIMENTS

We performed some experiments to get an impression of the applicability of our model. Further we compared the quality of our estimates with those based on the log-normal approximation [12].

First, we generated randomly a Bayesian belief network over 15 binary variables as in [3]. We did not use larger networks in order to be able to determine exact values of $G$. The probability table are selected once using a uniform distribution over the unit interval and once using a uniform distribution over the interval $[0, 0.1] \cup [0.9, 1]$. With this network, we generated a sample of 1000 cases using logic sampling [13]. This sample was used to estimate $\delta$ and $\alpha$ using Algorithm II with the median as robust function. An exact calculation of $G$ was performed by enumerating all instantiations of the 15 variables and counting corresponding probabilities. For comparison the accumulated probability estimated based on the log-normal approximation $\int_{-\infty}^{\log q} N(\mu, \sigma) / \int_{-\infty}^{0} N(\mu, \sigma)$ was calculated. The mean $\mu$ and variance $\sigma$ of the log-normal approximation were calculated exactly.

Figure 2 and 3 show the exact values of $G$ together with its approximations based on Formula 8 for network with probability tables selected from [0..1] with $u = 0.00005$ and from $[0, 0.1] \cup [0.9, 1]$ with $u = 0.001$, respectively. The value of $u$ was obtained by starting with $u = 0.005$ and consecutively lowering $u$ until the estimates of $\alpha$ and $\delta$ were based on about 50 cases. The y-axis shows the accumulated probability and the x-axis the values of $q$.

From the figure is clear that the approximation $\hat{G}$ differs only slightly from the exact values of $G$. So, the presented theory seems to give a good approximation. On the other hand, the approximation based on the log-normal distribution has a very large error; when applying this approximation we found that already at $q = 5.10^{-6}$ the error is larger than 0.9 for both cases rendering it inapplicable for estimating $G$.

We also performed experiments with a belief network with continuous variables. Let $V = \{X_1, X_2, X_3\}$ and consider the Bayesian belief network $(B_S, B_P)$ in Figure 4. Then, we have $P = f(x_1)f(x_2|x_1)f(x_3|x_1)$, so

$$P = \lambda \exp(-\lambda x_1) \exp(-x_1 x_2); \ x_1, x_2 > 0, 0 < x_3 < x_1. \quad (19)$$

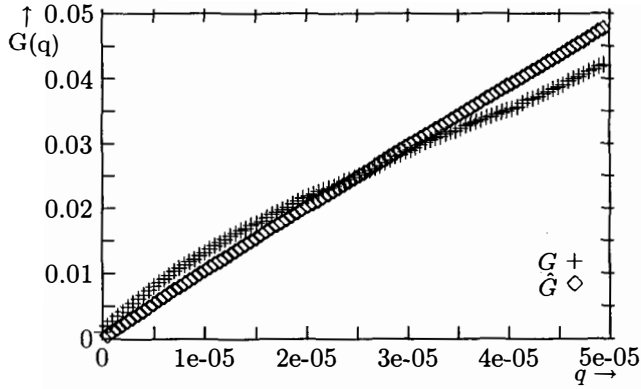

Figure 2: Accumulated probability, exact and approximate values with $u = 0.00005$ for network with probability tables selected from $[0..1]$.

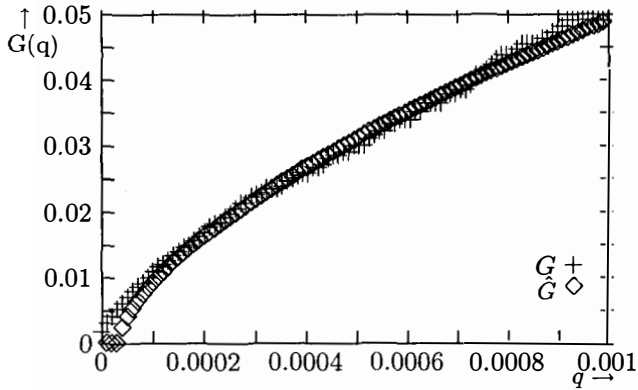

Figure 3: Accumulated probability, exact and approximate values with $u = 0.001$ for network with probability tables selected from $[0, 0.1] \cup [0.9, 1]$.

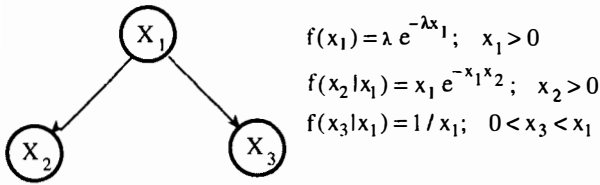

$$f(x_1) = \lambda \, e^{-\lambda x_1}; \quad x_1 > 0$$
$$f(x_2|x_1) = x_1 \, e^{-x_1 x_2}; \quad x_2 > 0$$
$$f(x_3|x_1) = 1/x_1; \quad 0 < x_3 < x_1$$

Figure 4: Example of a Bayesian belief network.

The exact distribution of the random variable $P$ can be calculated as follows:

- First we consider the change of variable:

$$\left. \begin{array}{l} P = \lambda \exp\left(-X_1(\lambda + X_2)\right) \\ U = X_1 \end{array} \right\} \quad (20)$$

- Next we obtain the joint density of $(P, U)$:

$$g(p, u) = 1; \ 0 < u < \frac{\log(\lambda/p)}{\lambda}; p > 0 \quad (21)$$

- Finally, we obtain the $P$-marginal:

$$f(p) = \frac{\log(\lambda/p)}{\lambda}; \ 0 \le p \le \lambda. \quad (22)$$

with cdf

$$F(p) = \begin{cases} 0 & if \quad p < 0 \\ \frac{p[1 - \log(p/\lambda)]}{\lambda} & if \quad 0 \le p \le \lambda \\ 1 & if \quad p > \lambda \end{cases} \quad (23)$$

Then

$$g(p) = pf(p) = \frac{p \log(\lambda/p)}{\lambda}; \ 0 \le p \le \lambda \quad (24)$$

and

$$G(p) = \frac{\int_0^p g(p)dp}{\int_0^\lambda g(p)dp} = \frac{p^2 \left(1 + 2\log(\lambda/p)\right)}{\lambda^2}; \ 0 \le p \le \lambda \quad (25)$$

Figure 5 shows $f(p)$ and the normalized $g(p)$, which is the pdf associated with $G(p)$, for $\lambda = 1$. From this we can find that 32% of the instantiations contribute 0.05% of the total probability.

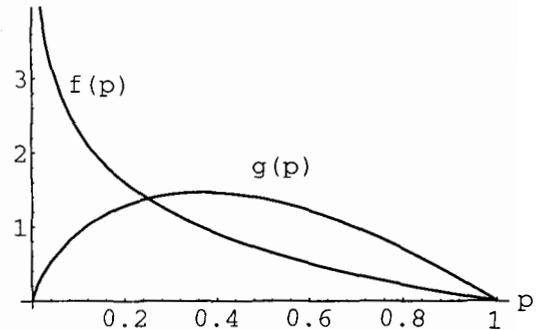

Figure 5: $f(p)$, and normalized $g(p)$ functions.

We can simulate $(X_1, X_2, X_3)$ using the conditional probabilities in Figure 4. We have simulated a sample of size 1000, which is shown in Figure 6.

The quality of the estimator (10) is illustrated in Figure 7, where the scatter plot of the estimated $G(p)$ and the exact $G(p)$ are shown.

We have selected a threshold value $u = 0.0951$, which corresponds to 0.05% of the total probability, and we have estimated $\delta$ and $\alpha$ using the Algorithm II. The obtained estimated are $\hat{\delta} = 0.0936$ and $\hat{\alpha} = 0.625$. Figure 8 shows the sample and the estimated model values, showing a good fitting. Finally, Figure 9 shows the exact and the estimated models for $G(p)$.

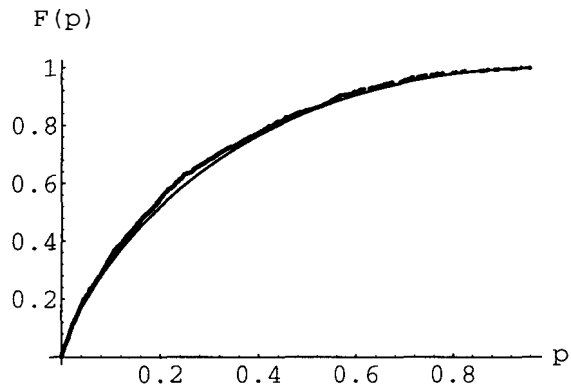

Figure 6: Exact, $F(p)$, and empirical cdfs.

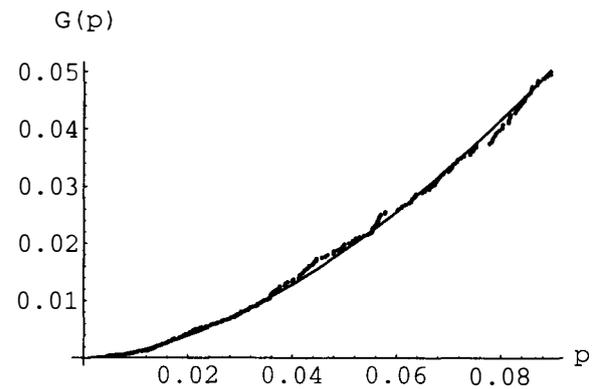

Figure 8: Sample and estimated model values below the threshold $u = 0.0951$

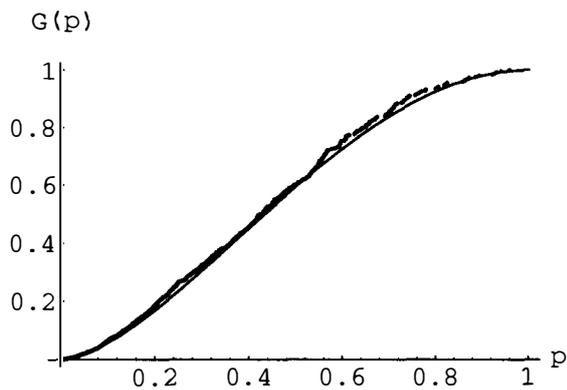

Figure 7: Exact $G(p)$ and estimated $\hat{G}(p)$ using (7).

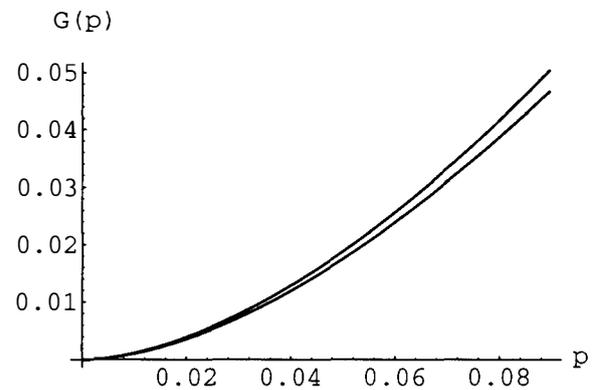

Figure 9: Exact and estimated models for $G(p)$ below threshold value $u = 0.0951$.

Table 1 shows the exact $F(p)$ and $G(p)$, and the estimated $\hat{G}(p)$ for different values of $p$.

Note that we used a Bayesian belief network with a few nodes only to be able to perform exact evaluation. However, there is just one random variable $P$ involved in the presented theory. So, the required sample size does not depend on the size of the network.

| $p$ | $F(p)$ | $G(p)$ | $\hat{G}(p)$ |
|-------|--------|-----------|-----------|
| 0.05  | 0.1998 | 0.0174    | 0.0176    |
| 0.01  | 0.0561 | 0.00102   | 0.00109   |
| 0.005 | 0.0315 | 0.000289  | 0.000268  |
| 0.002 | 0.0144 | 0.0000537 | 0.0000136 |

Table 1: Exact, $F(p)$ and $G(p)$ and estimated $\hat{G}(p)$ for different values of $p$.

## 6    CONCLUSIONS

We presented a method for estimating the total contribution $G(q)$ of the events with probability smaller than $q$ to the marginals based on extreme value theory. The estimate of $G(q)$ can be used in various techniques involving Bayesian belief networks: the number of iterations for some simulation algorithms [3, 4] can be calculated such that a certain error bound is guaranteed; the number of assignments in discrete approximation algorithms [5, 15] can be determined; a reasonable size of a cache [14] for storing cases can be computed.

Our theory is applicable to Bayesian belief networks with discrete, or continuous variables. The presented theory can easily be extended for belief networks with mixed discrete, and continuous variables. Experimental results suggest that our model gives a good approximation of $G$. However, the log-normal approximation proposed by Druzdzel [12] is inapplicable for estimating $G$.

It would be interesting to investigate techniques for in-crementally updating $\hat{G}$ when evidence is observed. If only discrete variables are involved, each instantiation can be stored together with case in the sample used for estimating $G$. When evidence is entered into the belief network, those cases in the sample that are not conform the observed evidence can be replaced by new to be generated cases. The other cases can stay unaltered in the sample, thus saving a lot of computational effort. Experiments have to be performed to get insight in how well the obtained sample is representative for $G$.

## APPENDIX: PROOF OF THEOREM

In this appendix we proof Theorem 2. Assume, without loss of generality that $i < j$, which implies $p_i \leq p_j$ and $C_i \leq C_j < 0$, and consider the following function of $\delta$,

$$h(\delta) = C_i \log(1 + (p_j - u)/\delta) - C_j \log(1 + (p_i - u)/\delta), \quad (26)$$

which is defined in the set $\{(-\infty, 0) \cup (u - p_i, \infty)\}$. Then, Equation (14) can be written as

$$h(\delta) = 0, \quad (27)$$

that is, the solutions of (14) are the zeroes of (26). Clearly, two zeroes of $h(\delta)$ are $\delta = \pm\infty$. We now show that there exist a finite solution of (27).

The function $h(\delta)$ has the following properties:

$$h(-\infty) = 0; \quad h(-0) = -\infty;$$

$$h(u - p_i) = -\infty; \quad h(\infty) = 0. \quad (28)$$

Additionally, it has no relative maximum and only one relative minimum which is given by

$$\frac{dh(\delta)}{d\delta} = \frac{1}{\delta}\left[\frac{C_i(u - p_j)}{\delta + p_j - u} - \frac{C_j(u - p_i)}{\delta + p_i - u}\right] = 0. \quad (29)$$

The solutions of (29) are $\delta_0 = \pm\infty$ and

$$\delta_0 = \frac{(C_i - C_j)(u - p_i)(u - p_j)}{C_i(u - p_j) - C_j(u - p_i)}. \quad (30)$$

Thus, we have, see Figure 10,

$$\begin{array}{lll} \delta_0 > 0 & \text{if} & u - p_i < C_i(u - p_j)/C_j, \\ \delta_0 \to \pm\infty & \text{if} & u - p_i \to C_i(u - p_j)/C_j, \\ \delta_0 < 0 & \text{if} & u - p_i > C_i(u - p_j)/C_j. \end{array} \quad (31)$$

The continuity of $h(\delta)$ together with (28) and the existence of a relative minimum, imply that $h(\delta)$ has only one finite zero if $C_j(u - p_i) \neq C_i(u - p_j)$. This zero is in the interval $(u - p_i, \delta_0)$ if $C_i(u - p_j) - C_j(u - p_i) <= 0$, see Figure 11, or in the interval $(\delta_0, 0)$ if $C_i(u - p_j) - C_j(u - p_i) > 0$, see Figure 12. Thus, we can use the bisection method to determine the solution of (14) (or the zero of $h(\delta)$). This completes the proof.

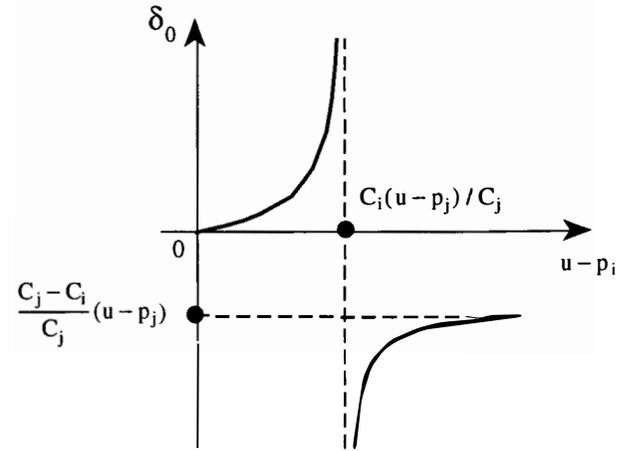

Figure 10: Plot of $\delta_0$ versus $u - p_i$ for a fixed value of $p_i \leq p_j$.

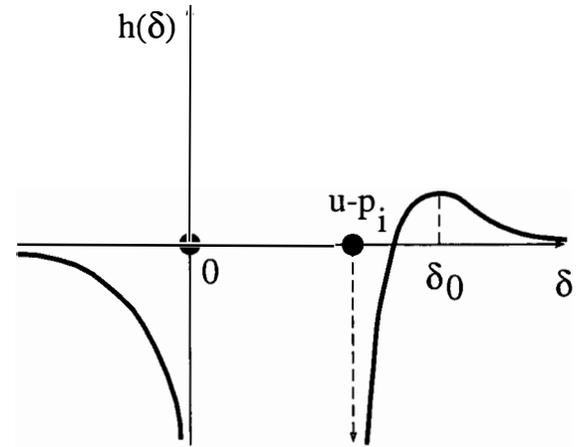

Figure 11: Plot of $h(\delta)$ versus $\delta$ for the case $u - p_i < C_i(u - p_j)/C_j$.

## Acknowledgements

The authors are grateful to the Dirección General de Investigación Científica y Técnica (DGICYT) (project PB92- 0504), for partial support of this work. The second author likes to express his thanks to Enrique Castillo and José Manuel Gutiérrez for inviting him to the University of Cantabria.

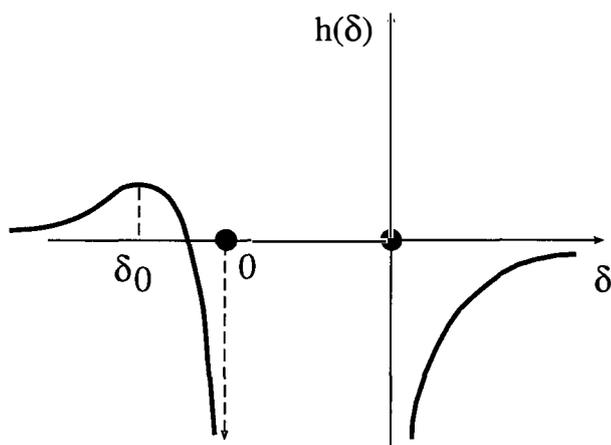

Figure 12: Plot of $h(\delta)$ versus $\delta$ for the case $u - p_i > C_i(u - p_n)/C_j$.